\documentclass[letterpaper, 10 pt, journal, twoside]{IEEEtran}

\IEEEoverridecommandlockouts                              

\usepackage{amsmath} 
\usepackage{amssymb}  
\usepackage{tensor}
\usepackage{mathtools}
\usepackage[binary-units=true]{siunitx}
\usepackage{booktabs}

\title{Robust Legged Robot State Estimation \\ Using Factor Graph Optimization}

\author{David Wisth, Marco Camurri, and Maurice Fallon%
\thanks{Manuscript received: February, 24, 2019; Revised June, 13, 2019;
Accepted July, 20, 2019.}%
\thanks{This paper was recommended for publication by Editor Nikos Tsagarakis
upon evaluation of the Associate Editor and Reviewers' comments. This research
has been conducted as part of the ANYbotics research community. It was part
funded by the Innovate UK-funded ORCA Robotics Hub (EP/R026173/1), the EU H2020
Project THING, a Royal Society University Research
Fellowship (Fallon) and a Google DeepMind studentship (Wisth).}%
\thanks{The authors are with the Oxford Robotics Institute at the University
of Oxford, United Kingdom
\texttt{\{davidw, mcamurri, mfallon\}@robots.ox.ac.uk}}%
\thanks{Digital Object Identifier (DOI): see top of this page.}
}

\usepackage{amssymb} 
\usepackage{algorithm2e} 
\usepackage{multirow} 

\usepackage{tikz}
\usetikzlibrary{bayesnet}
\usetikzlibrary{arrows,positioning}

\newcommand{\X}{\mathcal{X}}
\newcommand{\Z}{\mathcal{Z}}
\DeclareMathOperator*{\argmax}{arg\,max}
\DeclareMathOperator*{\argmin}{arg\,min}

\newcommand{\hide}[1]{}
\newcommand{\Figure}{Fig.~}

\newcommand{\ie}{{i.e.,~}}
\newcommand{\eg}{{e.g.,~}}

\newcommand{\bdmath}{\begin{dmath}}
\newcommand{\edmath}{\end{dmath}}
\newcommand{\beq}{\begin{equation}}
\newcommand{\eeq}{\end{equation}}
\newcommand{\bdm}{\begin{displaymath}}
\newcommand{\edm}{\end{displaymath}}
\newcommand{\bea}{\begin{eqnarray}}
\newcommand{\eea}{\end{eqnarray}}
\newcommand{\beal}{\beq \begin{array}{ll}}
\newcommand{\eeal}{\end{array} \eeq}
\newcommand{\beas}{\begin{eqnarray*}}
\newcommand{\eeas}{\end{eqnarray*}}
\newcommand{\ba}{\begin{array}}
\newcommand{\ea}{\end{array}}
\newcommand{\bit}{\begin{itemize}}
\newcommand{\eit}{\end{itemize}}
\newcommand{\ben}{\begin{enumerate}}
\newcommand{\een}{\end{enumerate}}

\newcommand{\Real}{\mathbb{R}}

\newcommand{\SEthree}{\ensuremath{\mathrm{SE}(3)}\xspace}

\newcommand{\calC}{{\cal C}}

\newcommand{\calI}{{\cal I}}

\newcommand{\calQ}{{\cal Q}}

\newcommand{\calT}{{\cal T}}

\newcommand{\T}{\mathbf{T}}
\newcommand{\R}{\mathbf{R}}

\newcommand{\transpose}{\mathsf{T}}

\newcommand{\tran}{\mathbf{p}}

\newcommand{\vel}{\mathbf{v}}

\newcommand{\bias}{\mathbf{b}}

\newcommand{\gravity}{\mathbf{g}}

\newcommand{\World}{\mathtt{W}}
\newcommand{\Imu}{\mathtt{I}}
\newcommand{\Camera}{\mathtt{C}}
\newcommand{\world}{\mathtt{{W}}}

\newcommand{\base}{\mathtt{{B}}}
\newcommand{\Base}{\mathtt{{B}}}

\newcommand{\foot}{\mathtt{K}}

\makeatletter
\let\NAT@parse\undefined
\makeatother
\usepackage{hyperref}
\begin{document}


\onecolumn
\thispagestyle{empty}

\hspace{3cm}
\begin{center}
This paper has been accepted for publication in \emph{IEEE Robotics And Automation Letters} (RA-L).\\

\hspace{1cm}

DOI: \href{http://dx.doi.org/10.1109/LRA.2019.2933768}{10.1109/LRA.2019.2933768}\\

IEEE Explore: \url{https://ieeexplore.ieee.org/document/8790726} \\

\hspace{1cm}

Please cite the paper as: \\

\hspace{1cm}

David Wisth, Marco Camurri, Maurice Fallon, \\
``Robust Legged Robot State Estimation Using Factor Graph Optimization'', \\
in \emph{IEEE Robotics And Automation Letters}, 2019.\\

\end{center}
\twocolumn


\setcounter{page}{1}

\maketitle

\begin{abstract}
Legged robots, specifically quadrupeds, are becoming increasingly attractive for
industrial applications such as inspection.
However, to leave the laboratory and to become useful to an end user requires reliability
in harsh conditions. From the perspective of state estimation, it is essential
to be able
to accurately estimate the robot's state despite challenges such as
uneven or slippery terrain, textureless and reflective scenes, as well as
dynamic camera occlusions.
We are motivated to reduce the dependency
on foot contact classifications, which fail when slipping, and to reduce position drift
during dynamic motions such as trotting.
To this end, we present a factor graph optimization method for state estimation
which tightly fuses and smooths inertial navigation, leg odometry and visual odometry.
The effectiveness of the approach is demonstrated using the ANYmal
quadruped robot navigating in a realistic outdoor industrial environment. This experiment included
trotting, walking, crossing obstacles and ascending a staircase. The proposed
approach decreased the relative position error by up to 55\% and
absolute position error by 76\% compared to
kinematic-inertial odometry.
\end{abstract}


\section{Introduction}
\label{sec:introduction}

\IEEEPARstart{F}{or} legged robots to become truly autonomous and useful they
must have a
consistent and accurate understanding of their location in the world. This is
essential for almost every aspect of robot navigation, including
control, motion generation, path planning, and local mapping.

Legged robots pose unique challenges to state estimation. First, the dynamic
motions generated by the robot footsteps can induce motion blur on camera
images as well as slippage or flexibility in the kinematics. Second, the strict
real-time requirements of legged locomotion require low latency, high frequency estimates
which are robust.
Third, the sensor messages are heterogeneous, with different frequencies and
latencies.
Finally, the conditions where legged robots are expected to operate are far
from ideal: poorly lit or textureless areas, self-similar structures, muddy or
slippery ground are some examples.

For these reasons, legged robots have traditionally relied on
filter-based state estimation, using \textit{proprioceptive} inputs
(IMUs, force/torque sensors and joint encoders) \cite{Bloesch2012, Camurri2017,
Bloesch2017b}.
While these approaches
give reliable and high frequency estimates, they are limited in their ability
to reject linear and angular position drift.

\begin{figure}[t]
\vspace{3mm}
\centering
\includegraphics[width=\columnwidth]{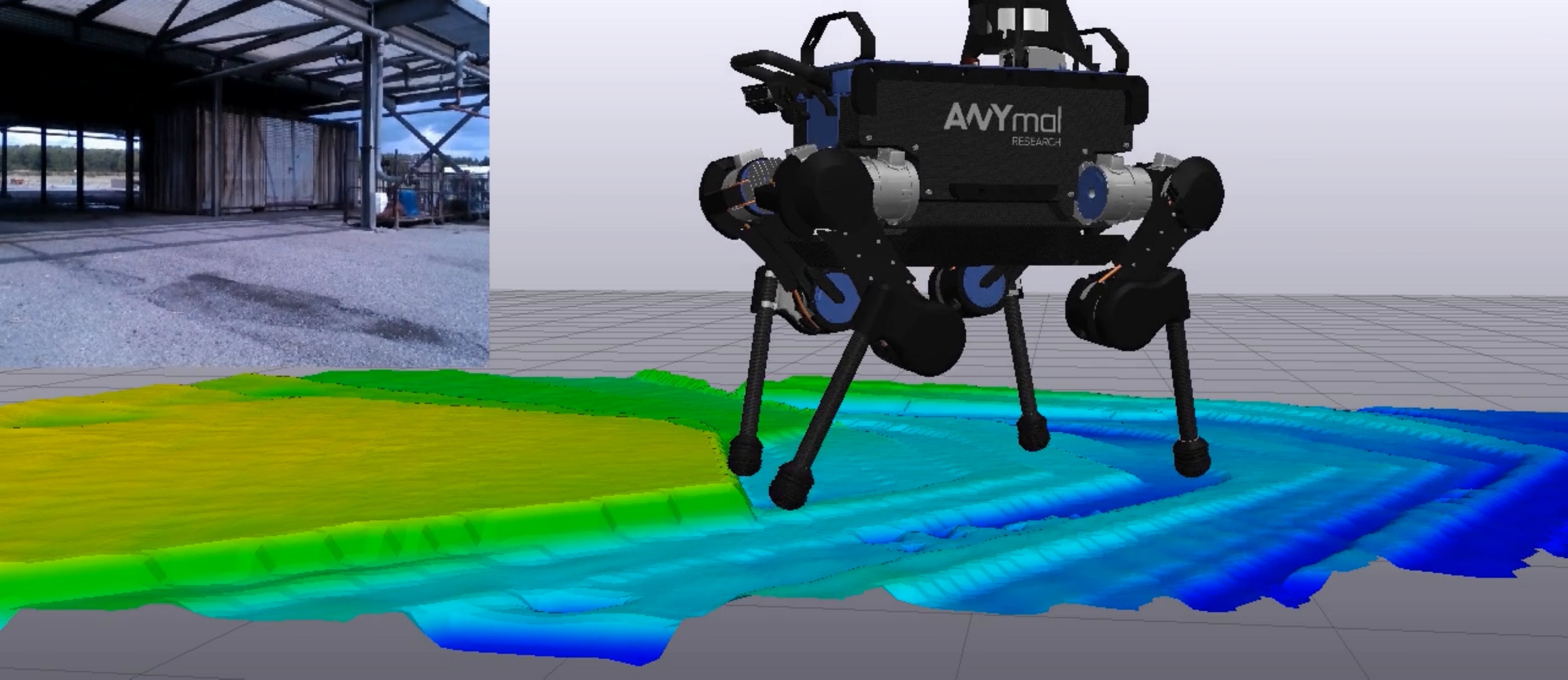}
\caption{An elevation map created by the GridMap package \cite{Hutter2017b}
while walking over flat terrain shows discrete ridges due to position drift in the
current kinematic state estimator. This makes dynamic locomotion and footstep planning
much more challenging.}
\vspace{-3mm}
\label{fig:elevation-map}
\end{figure}

For statically stable walking, leg odometry drift is low enough that terrain
mapping can be used
for continuous footstep planning and execution
\cite{fankhauser2018icra}. However for dynamic locomotion, position drift is
much higher which makes
such mapping ineffective, as illustrated in \Figure \ref{fig:elevation-map}.
The result of this is that we do not know what the shape of the terrain is under the robot due
to drift and cannot properly plan motions.

\begin{figure}[t]
\centering
\includegraphics[height=5cm]{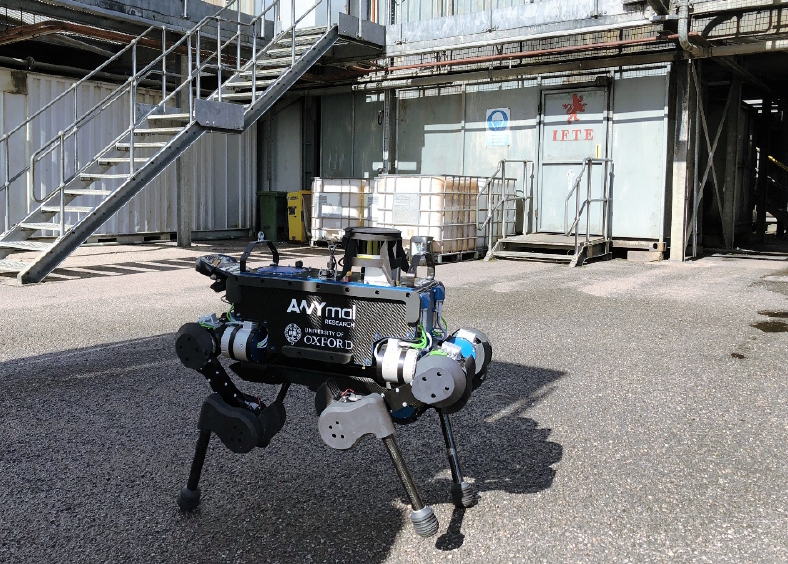}
\caption{Experiments were conducted using an ANYbotics ANYmal quadruped \cite{Hutter2016} in
outdoor environments, including the Oil Rig training facility shown above.
The ANYmal robot has 12 actuated degrees of freedom, an IMU, dual forward-facing
RealSense D435 cameras and a Velodyne VLP-16 LIDAR. Video:
\texttt{\url{https://youtu.be/p8o7mJPy4_w}.}}
\label{fig:anymal-fsc}
\end{figure}

To overcome this limitation, some previous works have
incorporated \textit{exteroceptive}
inputs (cameras and LIDAR) into
filtering estimators in a loosely coupled fashion. This has been
successfully demonstrated on legged machines operating in field experiments
\cite{Ma2016, Nobili2017}.

However, because these filters marginalize all previous states of the robot,
it is not possible to fully exploit a (recent) history of measurements, as smoothing
methods can.

Research into smoothing approaches applied to Visual-Inertial Navigation Systems
(VINS) is now well established in the Micro-Aerial-Vehicle (MAV) community.
On a MAV this approach has been successful due to careful time
synchronization of the IMU and cameras, smooth vehicle motion, and also the
absence of the challenges of articulated legged machines.

A recent work by Hartley et~al. \cite{Hartley2018a}
demonstrated that a VINS approach could be adapted to
legged robots --- in their case, a biped. Their initial results were promising,
but only tested in a controlled scenario. Additionally,
vision was integrated as relative pose constraints, rather than incorporating
the feature residuals directly into the optimization.

\subsection*{Contribution}
This paper aims to progress the deployment of state
estimation smoothing methods in realistic application scenarios.
Compared to previous research, we present the following contributions:

\begin{itemize}
\item We present the first state estimation method based on factor graphs that
tightly integrates visual features directly into the cost function (rather than adding a pose
constraint from a separate visual inertial module), together with preintegrated
inertial factors and relative pose constraints from the kinematics. We will
refer to our proposed method as VILENS (Visual Inertial LEgged Navigation System);

\item We demonstrate the performance and robustness of our method with extensive
experiments on two field scenarios. Challenges included motion blur, dynamic scene
occludants, textureless and reflective scenes, and locomotion on uneven, muddy
and slippery
terrain;

\item We demonstrate that a low-cost consumer-grade depth camera, the RealSense D435, is sufficient
to significantly improve the state estimate in these conditions.
\end{itemize}

The remainder of the article is presented as follows: in Section
\ref{sec:related-work} we describe the previous research in the field; the
theoretical background of the algorithm
is described in Sections \ref{sec:problem-statement} and
\ref{sec:factor-definitions}. Section \ref{sec:implementation} outlines the
details of our implementation. The experimental results and their
discussion are shown in Sections \ref{sec:experimental-results} and
\ref{sec:discussion-and-limitations}. Finally, Section
\ref{sec:conclusions} concludes the article.

\section{Related Work}
\label{sec:related-work}
Multi-sensor fusion for mobile robot state estimation has been widely described
in the literature \cite{Barfoot2017}.
Here, we limit our discussion to filtering and smoothing approaches with particular
focus on dynamic legged robots.

\subsection{Filtering Approaches}
After the spread of filtering methods for proprioceptive state estimation
\cite{Bloesch2012}, there has been interest in including exteroceptive data,
particularly Visual Odometry (VO).

Ma et al. \cite{Ma2016} presented a method based on an
Extended Kalman Filter (EKF) with an error state formulation developed for the
Boston Dynamics LS3. The system was
primarily driven by inertial predictions with VO updates: a modular sensor head
performs the fusion of a very high quality tactical grade IMU with two hardware
synchronized stereo cameras. The leg odometry (fused with an additional
navigation grade IMU in the body) was used only in case of VO
failure. Their extensive evaluation (over several \si{\kilo\meter}) achieved
\SI{1}{\percent} error per distance traveled.

Nobili et al. \cite{Nobili2017} recently presented a state estimator for the HyQ
quadruped robot which used an EKF to combine inertial,
kinematic, visual, and LIDAR measurements. In contrast to \cite{Ma2016}, the EKF
was driven by an inertial process model with the primary corrections coming from leg odometry (synchronized
in EtherCAT) at the nominal control rate (\SI{1}{\kilo\hertz}). The VO
updates and an ICP-based matching algorithm were run on a separate computer at
lower frequency and integrated into the estimator when available. This allowed
the use of the estimator inside the control loop and has been recently
demonstrated in dynamic motions with local mapping \cite{villarreal2019ral}.

\subsection{Smoothing Approaches}
There has been a significant body of work on visual inertial navigation, especially for use
with MAVs. A recent benchmark paper \cite{Delmerico2018} evaluated a number of
state-of-the-art methods on the EuRoC dataset \cite{Burri2016}. The maturity of
the field was highlighted by the fact that many algorithms achieved an average
Relative Position Error (RPE) of less than \SI{20}{\centi\meter} per
\SI{35}{\meter} traveled. The authors concluded that the best performing
algorithms were OKVIS \cite{Leutenegger2015}, ROVIO \cite{Bloesch2015a}, and
VINS-Mono \cite{Qin2017}. All of these algorithms perform windowed
optimization to achieve the most accurate state estimate while bounding
computation time. An exception was SVO+GTSAM \cite{Forster2017} which loosely
coupled the SVO visual odometry algorithm with IMU data using iSAM2
as the smoothing back-end \cite{Kaess2012}.

The methods were, however, typically designed assuming that the IMU and cameras
were synchronized in hardware, and that the vehicle motions were smooth. They are
difficult to implement on legged platforms due to their hardware complexity and
the high vibrations caused by locomotion.

As mentioned above, the first approach to use smoothing/optimization on legged robots was the work of
Hartley et al. \cite{Hartley2018a} which presented a
fusion of kinematic, inertial, and visual information, again using iSAM2
as the smoothing back-end. They presented a
mathematical framework to model contact points as landmarks in the environment,
similar to \cite{Bloesch2012}. Their system incorporated contact information
from only a single contact point at a time, and directly integrates (as pose
constraints) the relative motion estimate of the SVO2
\cite{Forster2017a} algorithm. The approach was tested using inertial
and visual input from a MultiSense S7 sensor (which was hardware synchronized)
mounted on a Cassie biped (from Agility Robotics). A short indoor experiment
of \SI{60}{\second} showed that adding vision
into the optimization reduced the relative position error.

Our approach aims to combine best practice from the mature field of VINS --
including windowed optimization and tight integration of visual features -- with
legged odometry to provide robust state estimation for legged robots. In Section
\ref{sec:experimental-results} we show our system outperforming both
kinematic-inertial and visual-inertial approaches in large-scale outdoor urban and
industrial experiments.


\section{Problem Statement}
\label{sec:problem-statement}
We wish to track the linear position and velocity of a 12 Degrees of Freedom
(DoF) legged robot
equipped with an industrial grade MEMS IMU, an RGB/IR camera, joint encoders and
torque sensors. The sensor specifications are
detailed in Table \ref{tab:sensors}.

\begin{table}
\vspace{3mm}
\caption{Sensor specifications}
\centering
\begin{tabular}{lccl}
\toprule
\textbf{Sensor} & \textbf{Model} & \textbf{\si{\hertz}} & \textbf{Specs} \\
\midrule
\multirow{2}{*}{IMU} & \multirow{2}{*}{Xsens MTi-100} & \multirow{2}{*}{400} &
Init Bias: %
\SI{0.2}{\degree\per\second} $~\vert~$  \SI{5}{\milli\gram}  \\
 &  &  & Bias Stab: %
\SI{10}{\degree\per\hour} $~\vert~$  \SI{15}{\milli\gram}  \\
\midrule
\multirow{8}{*}{Camera} & \multirow{8}{*}{RealSense D435} & \multirow{8}{*}{30}
& \textit{Keble College Dataset:} \\
& & & Resolution: $848 \times 480$ px \\
& & &  FoV: \SI{91.2 x 65.5}{\degree} \\
& & & Imager: IR global shutter \\
& & & \textit{Oil Rig Dataset:} \\
& & & Resolution: $640 \times 480$ px \\
& & &  FoV: \SI{69.4 x 42.5}{\degree} \\
& & & Imager: RGB rolling shutter \\
\midrule
Encoder & ANYdrive & 400 & Resolution: \SI{<0.025}{\degree} \\
\midrule
Torque & ANYdrive & 400 & Resolution: \SI{<0.1}{\newton\meter} \\
\bottomrule
\end{tabular}
\label{tab:sensors}
\vspace{-3mm}
\end{table}

\begin{figure}
\vspace{3mm}
\centering
\includegraphics[height=5cm]{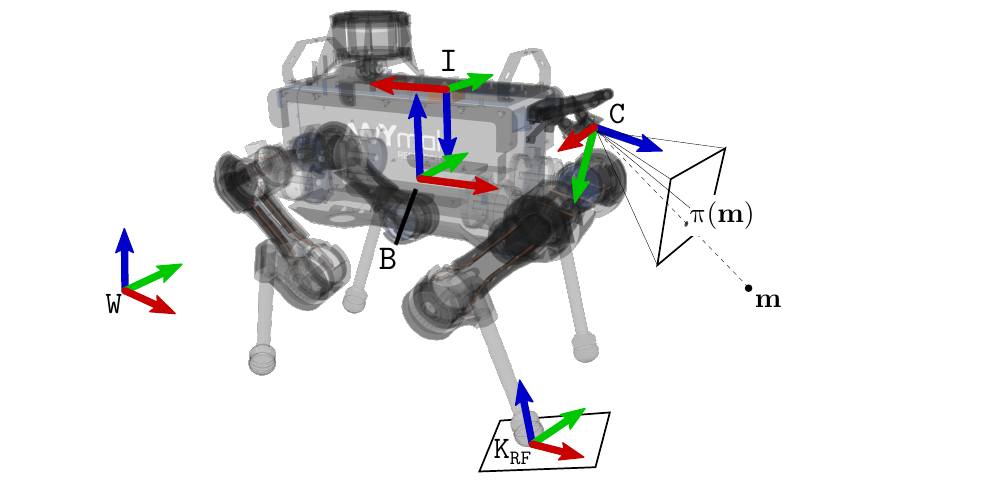}
\caption{Reference frames conventions. The world frame $\World$ is fixed to
earth, while the base frame $\Base$, the camera's
optical frame $\Camera$, and the IMU frame, $\Imu$ are rigidly attached to the
robot's chassis. When a foot touches the ground (\eg the Right Front, RF), a
contact frame $\foot$ (perpendicular to the ground and parallel to
$\World$'s $y$-axis) is defined. The projection of a landmark point
$\mathbf{m}$ onto the image plane is $\pi(\mathbf{m})$.}
\label{fig:coordinate-frames}
\vspace{-3mm}
\end{figure}

In \Figure \ref{fig:coordinate-frames} we provide a schematic of the reference
frames involved. The pose of the robot's base $\Base$ expressed in the fixed-world, inertial frame
$\World$ is defined as:
\[\T_{\World\Base} =
\begin{bmatrix}
\R_{\world\base} & \tensor[_\world]{\tran}{_{\world\base}} \\
\boldsymbol{0} & 1%
\end{bmatrix} \in \SEthree
\]
The IMU and camera sensing frames are $\Imu$ and $\Camera$,
respectively. The relative transformations between $\Base$, $\Imu$ and $\Camera$
are assumed to be known from CAD design. The location of a foot in base
coordinates is expressed as $\tensor[_\base]{\mathbf{p}}{_{\base\foot}}$.

\subsection{State Definition}
Borrowing the notation from \cite{Forster2017}, we define the state of the
system at time $t_i$ as:
\beq
  \boldsymbol{x}_i \triangleq [\R_i,\tran_i,\vel_i,\bias_i]
\eeq
where the couple $(\R_i, \tran_i)$ expresses the robot pose and
$\vel_i \in \Real^3$ is the robot linear velocity. As is common in this field, the
stack
of gyro and accelerometer IMU
biases $\bias_i = [\bias^\omega_i \;\;
\bias^a_i] \in \Real^6$ replaces the angular velocity, which is directly
measured by the IMU.

Let $\mathsf{K}_k$ be the set of camera keyframe indices up to time
$t_k \in \Real$. We assume that for each keyframe image
$\calC_i$ (with $i\in\mathsf{K_k}$) a number of landmark points
$\tensor[_\world]{\mathbf{m}}{_j}$ are visible, where $j \in \mathsf{M}_i
\subseteq
\mathsf{M}$; $\mathsf{M}_i$ indicates the set of landmark indices visible from
keyframe $\calC_i$ out of the full set of landmarks, $\mathsf{M}$.
We then define the objective of our estimation problem $\X_k$ as the history of
robot states and landmarks
detected up to $t_k$:
\begin{equation}
\X_k \triangleq \bigcup_{\forall i \in
\mathsf{K}_k} \left[\{\boldsymbol{x}_i\}_,
\bigcup_{\forall j \in \mathsf{M}_i}\{\mathbf{m}_j\}\right]
\label{eq:state-history}
\end{equation}

\subsection{Measurements}
\begin{figure}
 \centering
 \includegraphics[width=\columnwidth]{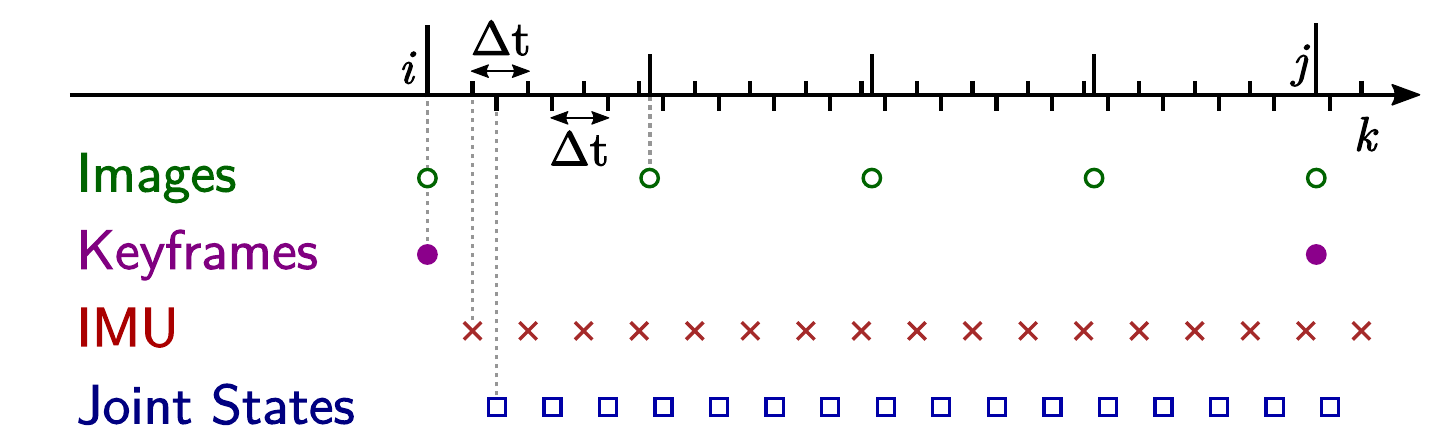}
 \caption{The sensors inputs are images, IMU measurements, and joint
 states, which are in general unsynchronized with each other.}
 \label{fig:sensor-inputs}
\end{figure}

The input measurements consist of camera images, IMU readings, and joint sensing
(position, velocity and torque). The measurements are not assumed to be
synchronized. However, we assume they
have a common time frame. The IMU and joint states have the same frequency
(see \Figure \ref{fig:sensor-inputs}).
For each pair of consecutive keyframe indices $\Delta i = i-1,
i \in
\mathsf{K}_k$ we define $\mathsf{I}_{\Delta i}$  as the set of IMU measurements
indices such that $\forall~m \in
\mathsf{I}_{\Delta i}$ we have $t_{i-1} \le t_m < t_{i}$. We
then indicate with $\calI_{\Delta i} = \bigcup_{\forall j \in
\mathsf{I}_{\Delta i}}(\boldsymbol{\omega}_j,{\mathbf{a}}_j)$ all angular
velocity and proper acceleration measurements collected between time $t_{i-1}$
and
$t_{i}$. Analogous definitions apply to joint states $\calQ_{\Delta i} =
\bigcup_{\forall j \in \mathsf{Q}_{\Delta i}}
(\mathbf{q}_j,\dot{\mathbf{q}}_j, \boldsymbol{\tau}_j)$, which include all
joint
positions, velocities and torques collected
between
time $t_{i-1}$ and $t_{i}$.
Strategies to account for synchronization issues are discussed in Section
\ref{sec:synch}.

Finally, we then let $\Z_k$ denote the set of all measurements up to time $t_k$:
\begin{equation}
\Z_k \triangleq \bigcup_{\forall i \in \mathsf{K}_k} \left\{ \calI_{\Delta i},
\calC_{i}, \calQ_{\Delta i} \right\}
\label{eq:measurement}
\end{equation}

\subsection{Maximum-A-Posteriori Estimation}

The aim of the factor graph framework is to maximize the posterior the history
of states and landmarks $\X_k$, given the history of measurements $\Z_k$:
\begin{equation}
 \X^*_k = \argmax_{\X_k} p(\X_k|\Z_k) \propto
p(\X_0)p(\Z_k|\X_k)
\label{eq:posterior}
\end{equation}
Where the last member of (\ref{eq:posterior}) is the likelihood function, which
is
proportional to the posterior and therefore can be used as a cost function.
If the measurements
are conditionally independent
and corrupted by zero mean Gaussian noise, then (\ref{eq:posterior})
is equivalent to a least squares problem of the form:
\begin{equation}
  \X^* = \argmin_{\X_k} \sum_{\calT} \sum_{\forall i \in
\mathsf{K}_k} \| \mathbf{r}_{\calT_i} \|^2_{\Sigma_{\calT_i}}
\label{eq:cost-function}
\end{equation}
where $\mathbf{r}_{\calT_i}$ is the residual of the error between the predicted
and measured value of type $\calT$ (\eg IMU $\calI$) at keyframe index
$i \in \mathsf{K}_k$. The
quadratic cost of each residual is weighted by the corresponding covariance
$\Sigma_{\calT_i}$.

\begin{figure}
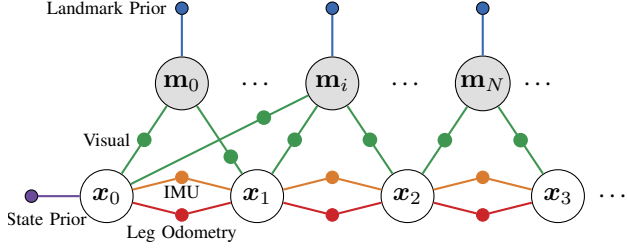

	\begin{center}
		\tikz{ %
            \clip(-4.3,-0.6) rectangle (4,3);
			\definecolor{color1}{RGB}{57,106,177};
			\definecolor{color2}{RGB}{218,124,48};
			\definecolor{color3}{RGB}{204,37,41};
			\definecolor{color4}{RGB}{62,150,81};
			\definecolor{color5}{RGB}{107,76,154};
			%
			\node[latent, xshift=-3cm,yshift=0cm] (X_0) {$\boldsymbol{x}_0$};
			\node[latent, xshift=-1cm,yshift=0cm] (X_1) {$\boldsymbol{x}_1$};
			\node[latent, xshift= 1cm,yshift=0cm] (X_2) {$\boldsymbol{x}_2$};
			\node[latent, xshift= 3cm,yshift=0cm] (X_3) {$\boldsymbol{x}_3$};
			\node[latent, xshift= 4.5cm,yshift=0cm, draw=none, fill=none] (X_4)
			{};
			%
			\node[obs, xshift=-2cm,yshift=1.5cm]  (L_0) {$\mathbf{m}_0$};
			\node[obs, xshift=0cm, yshift=1.5cm]  (L_i) {$\mathbf{m}_i$};
			\node[obs, xshift=2cm, yshift=1.5cm]  (L_N) {$\mathbf{m}_N$};
			\node[obs, xshift=3.5cm, yshift=1.5cm, draw=none, fill=none] (L_Np1)
		    {};
			%
			\path (L_0) -- node[auto=false]{\ldots} (L_i);
			\path (L_i) -- node[auto=false]{\ldots} (L_N);
			\path (X_3) -- node[auto=false]{\ldots} (X_4);
			\path (L_N) -- node[auto=false]{\ldots} (L_Np1);
			%
			\node[midway,circle,draw,fill=color1,xshift=-2cm,yshift=2.5cm,scale=0.5]
			 (factor_L0) {};
			\node[midway,circle,draw,fill=color1,xshift=0cm,yshift=2.5cm,scale=0.5]
			(factor_Li) {};
			\node[midway,circle,draw,fill=color1,xshift=2cm,yshift=2.5cm,scale=0.5]
			(factor_LN) {};
			\draw[thick, color1] (L_0) -- (factor_L0);
			\draw[thick, color1] (L_i) -- (factor_Li);
			\draw[thick, color1] (L_N) -- (factor_LN);
			\node[midway,circle,draw,fill=none,draw=none,xshift=-3cm,yshift=2.5cm,scale=0.7]
			(label_imu) {Landmark Prior};
			%
			\node[midway,circle,draw,fill=color5,xshift=-4cm,yshift=0cm,scale=0.5]
			(factor_X0) {};
			\draw[thick, color5] (X_0) -- (factor_X0);
			\node[midway,circle,draw,fill=none,draw=none,xshift=-3.8cm,yshift=-0.3cm,scale=0.7]
			(label_imu) {State Prior};
			%
			\node[midway,circle,draw,fill=color2,draw=color2,xshift=-2cm,yshift=0.25cm,scale=0.5]
			(factor_imu0) {};
			\node[midway,circle,draw,fill=color2,draw=color2,xshift=0cm,yshift=0.25cm,scale=0.5]
			(factor_imu1) {};
			\node[midway,circle,draw,fill=color2,draw=color2,xshift=2cm,yshift=0.25cm,scale=0.5]
			(factor_imu2) {};
			\draw[thick, color2] (X_0) -- (factor_imu0);
			\draw[thick, color2] (X_1) -- (factor_imu0);
			\draw[thick, color2] (X_1) -- (factor_imu1);
			\draw[thick, color2] (X_2) -- (factor_imu1);
			\draw[thick, color2] (X_2) -- (factor_imu2);
			\draw[thick, color2] (X_3) -- (factor_imu2);
			\node[midway,circle,draw,fill=none,draw=none,xshift=-2cm,yshift=0.05cm,scale=0.7]
			(label_imu) {IMU};
			%
			\node[midway,circle,draw,fill=color3,draw=color3,xshift=-2cm,yshift=-0.25cm,scale=0.5]
			(factor_lo0) {};
			\node[midway,circle,draw,fill=color3,draw=color3,xshift=0cm,yshift=-0.25cm,scale=0.5]
			(factor_lo1) {};
			\node[midway,circle,draw,fill=color3,draw=color3,xshift=2cm,yshift=-0.25cm,scale=0.5]
			(factor_lo2) {};
			\draw[thick, color3] (X_0) -- (factor_lo0);
			\draw[thick, color3] (X_1) -- (factor_lo0);
			\draw[thick, color3] (X_1) -- (factor_lo1);
			\draw[thick, color3] (X_2) -- (factor_lo1);
			\draw[thick, color3] (X_2) -- (factor_lo2);
			\draw[thick, color3] (X_3) -- (factor_lo2);
			\node[midway,circle,draw,fill=none,draw=none,xshift=-2cm,yshift=-0.5cm,scale=0.7]
			(label_leg) {Leg Odometry};
			%
			\draw [thick, color4] (X_0) -- +(L_0)
			node[midway,circle,draw,fill=color4,scale=0.5] {};
			\draw [thick, color4] (X_0) -- +(L_i)
			node[near end,circle,draw,fill=color4,scale=0.5] {};
			\draw [thick, color4] (X_1) -- +(L_0)
			node[near start,circle,draw,fill=color4,scale=0.5] {};
			\draw [thick, color4] (X_1) -- +(L_i)
			node[midway,circle,draw,fill=color4,scale=0.5] {};
			\draw [thick, color4] (X_2) -- +(L_i)
			node[midway,circle,draw,fill=color4,scale=0.5] {};
			\draw [thick, color4] (X_2) -- +(L_N)
			node[midway,circle,draw,fill=color4,scale=0.5] {};
			\draw [thick, color4] (X_3) -- +(L_N)
			node[midway,circle,draw,fill=color4,scale=0.5] {};
			\node[midway,circle,draw,fill=none,draw=none,xshift=-3cm,yshift=0.75cm,scale=0.7]
			(label_visual) {Visual};
		}
		\caption{The factor graph consists of state and landmark nodes linked
		by prior, visual, inertial, and leg odometry factors.}
		\label{fig:factor-graph-structure}
	\end{center}
	\vspace{-3mm}
\end{figure}

From (\ref{eq:state-history}) and (\ref{eq:measurement}) the optimization
becomes the following:
\begin{multline}
  \X^{*} = \argmin_{\X} \sum_{j \in M}
  \|\mathbf{r}_{\mathbf{m}_{j,0}}\|^2_{\Sigma_{\mathbf{m}_{j,0}}} + \sum_{i \in
\mathsf{K}_k} \| \mathbf{r}_{\calI_{\Delta i}} \|^2_{\Sigma_{\calI_{\Delta i}}}
+ \\
+\|\mathbf{r}_0\|^2_{\Sigma_0} + \sum_{i \in \mathsf{K}_k} \sum_{j \in
\mathsf{M}_i}\|
\mathbf{r}_{\mathbf{m}_{j}} \|^2_{\Sigma_{\mathbf{m}_{j}}}
+ \sum_{i \in
\mathsf{K}_k} \| \mathbf{r}_{\calQ_{\Delta i}} \|^2_{\Sigma_{\calQ_{\Delta i}}}
\label{eq:final-cost}
\end{multline}
where the residuals are from:
landmark prior, IMU, state prior, camera and leg odometry factors,
respectively. These
factors will be used to create the
factor graph structure shown in \Figure \ref{fig:factor-graph-structure}.
In the next section we define each residual of (\ref{eq:final-cost}).

\section{Factor Definitions}
\label{sec:factor-definitions}

\subsection{Prior Factors}
Since this is an odometry system, prior factors are required during
initialization to anchor the unobservable modes (\ie position and
yaw) to a fixed reference frame. The residual is defined as the error between
the estimated state $\boldsymbol{x}_0$ and the prior
$\boldsymbol{x}_{p_0}$:
\begin{equation}
\mathbf{r}_{0}(\boldsymbol{x}_0, \Z) = \left(\begin{array}{c}
  \Phi (\T_0^{-1} \T_{p_0})\\
  \vel_0   - \vel_{p_0}   \\
  \bias_0^a - \bias_{p_0}^a \\
  \bias_0^\omega - \bias_{p_0}^\omega
  \end{array} \right)  \label{eq:prior-residual}
\end{equation}
where $\Phi : \SEthree \mapsto \Real^6$ is the lifting
operator \cite{Forster2017}. The prior state of the system is determined by
IMU initialization, assuming the robot is stationary.

\subsection{Visual Odometry Factors}
The visual odometry residual consists of two components. The first
is the difference between the measured landmark pixel location,
$(u_{i,j},\;v_{i,j})$, and the re-projection of the estimated landmark location
into image coordinates, $(\pi_u, \pi_v)$
using the standard radial-tangential
distortion model. The residual is defined as:
\begin{equation}
\mathbf{r}_{\mathbf{m}_{j}} =
\left( \begin{array}{c}
\pi_u(\R_i,\tran_i, \mathbf{m}_j) - u_{i,j} \\
\pi_v(\R_i,\tran_i, \mathbf{m}_j) - v_{i,j}
\end{array} \right)
\end{equation}

The second is the error between the prior on the landmark location
$\mathbf{m}_{j,0}$ and the estimated landmark location $\mathbf{m}_{j}$:
\begin{equation}
\mathbf{r}_{\mathbf{m}_{j,0}} = \mathbf{m}_{j} - \mathbf{m}_{j,0}
\end{equation}
Legged robots typically move dynamically with velocities parallel to the
optical axis of the camera. This motion can lead to underconstrained landmark
positions. To this end, a landmark prior is generated online through an
initial triangulation procedure (see Section \ref{sec:implementation}).
The covariance, $\Sigma_{\mathbf{m}_{i,j}}$, is determined by the
typical feature tracking accuracy (in our case, we use 1 pixel).

\subsection{Preintegrated IMU Factors}

We use the IMU preintegration algorithm described by Forster et~al.
\cite{Forster2017}. 
This approach preintegrates the IMU measurements between nodes in the factor graph
to provide high frequency state updates between optimization steps. The
preintegrated IMU measurements are then used to create a new IMU factor between
two consecutive keyframes.
This will use an error term of the form:
\begin{equation}
\mathbf{r}_{\calI_{\Delta i}}  = \left[ \mathbf{r}^\transpose_{\Delta
\R_{\Delta i}},
             \mathbf{r}^\transpose_{\Delta \tran_{\Delta i}},
             \mathbf{r}^\transpose_{\Delta \vel_{\Delta i}},
             \mathbf{r}^\transpose_{\Delta \bias_{\Delta i}} \right]
\end{equation}
where $\calI_{\Delta i}$ are the IMU measurements between times $i-1$ and $i$.
The individual elements of the residual are defined as:
\begin{eqnarray}
\mathbf{r}_{\Delta \R_{\Delta i}}   & = & \log\left( \Delta\tilde{\R}_{\Delta
i}(b^g_{i-1}) \right) \R^T_{i-1} \R_i\\
\mathbf{r}_{\Delta \tran_{\Delta i}} & = & \R^T_{i-1} \left( \tran_i -
\tran_{i-1} - \vel_{i-1}\Delta t_{\Delta i} - \frac{1}{2}\gravity \Delta
t_{\Delta i}^2 \right) \nonumber\\
& & - \Delta \tilde{\tran}_{\Delta i}(b^g_{i-1}, b^a_{i-1}) \\
\mathbf{r}_{\Delta \vel_{\Delta i}}  & = &  \R^T_{i-1} \left( \vel_i -
\vel_{i-1} - \gravity \Delta t_{\Delta i} \right) \nonumber\\
& & - \Delta \tilde{\vel}_{\Delta i}(b^g_{i-1}, b^a_{i-1}) \\
\mathbf{r}_{\Delta \bias_{\Delta i}}  & = & \bias_{i} - \bias_{i-1}
\end{eqnarray}
where $\Delta\tilde{\R}_{\Delta i}, \Delta \tilde{\tran}_{\Delta i}, \Delta
\tilde{\vel}_{\Delta i} $ are the preintegrated IMU measurements defined in
\cite{Forster2017}. The covariance, $\Sigma_{\calI_{\Delta i}}$, is based on an
offline Allan Variance analysis.

\subsection{Leg Odometry Factors}

Leg Odometry (LO) estimates the incremental motion of a
walking robot given its kinematic sensing and the perceived
contacts between the robot's feet and the ground. The main assumption behind LO
measurements is that the
absolute velocity of a contact point is zero. This condition is met when
the Ground Reaction Force (GRF) at a contact point is inside a hypothetical
friction
cone. Torques at contact points are neglected as quadruped robots' feet are
idealized as
points.

Since the GRF, the terrain inclination, and the coefficient of static friction
are typically unknown, an appropriate fusion of the dynamic model,
kinematics, and IMU is required. In this work, we rely on the state estimator
provided by ANYbotics, the Two-State Implicit Filter (TSIF) \cite{Bloesch2017b}.
We use the relative robot motion estimated by the TSIF to formulate relative
pose factors so as to further constrain the robot's motion estimate from the
factor graph.

Given two keyframes at times $t_{i-1}$ and $t_{i}$, the corresponding pose
estimates
$\widetilde{\T}_{i-1},\;\widetilde{\T}_{i}$ are computed via linear/slerp
interpolation from the filter history and used to produce the relative pose
constraint:
\begin{equation}
 \mathbf{r}_{\calQ_{\Delta i}} = \Phi\left((\T_{i-1}^{-1}\T_{i})^{-1}
\widetilde{\T}_{i-1}^{-1}\widetilde{\T}_{i} \right)
\label{eq:relative-pose-factor}
\end{equation}
where the covariance $\Sigma_{\calQ_{\Delta i}}$ is provided by the filter, and
$\Phi$ is the lifting operator defined in \cite{Forster2017}.

\subsection{Zero Velocity Update Factors}
To limit drift and factor graph growth when the robot is stationary, we detect
zero velocity motion from camera frames
by calculating the average feature motion:
\begin{equation}
\overline{\Delta x} = \frac{\sum_{i=k-N}^{k} \left( \sqrt{(u_i - u_{i-1})^2 +
(v_i - v_{i-1})^2} \right)}{N+1}
\end{equation}
If $\overline{\Delta x}$ is below a certain threshold $\beta=0.5$ pixels over
$N=10$ successive frames, we
stop adding image landmark measurements to the graph and simply add a zero
relative pose factor to the graph of the same form as
(\ref{eq:relative-pose-factor}). The thresholds were tuned by calculating
$\overline{\Delta x}$ when the camera was stationary. These factors reduce
computation when the robot is stationary and can reduce overall drift by up to
20\%.


\section{Implementation}
\label{sec:implementation}

\begin{figure}
\vspace{3mm}
\includegraphics[width=\columnwidth]{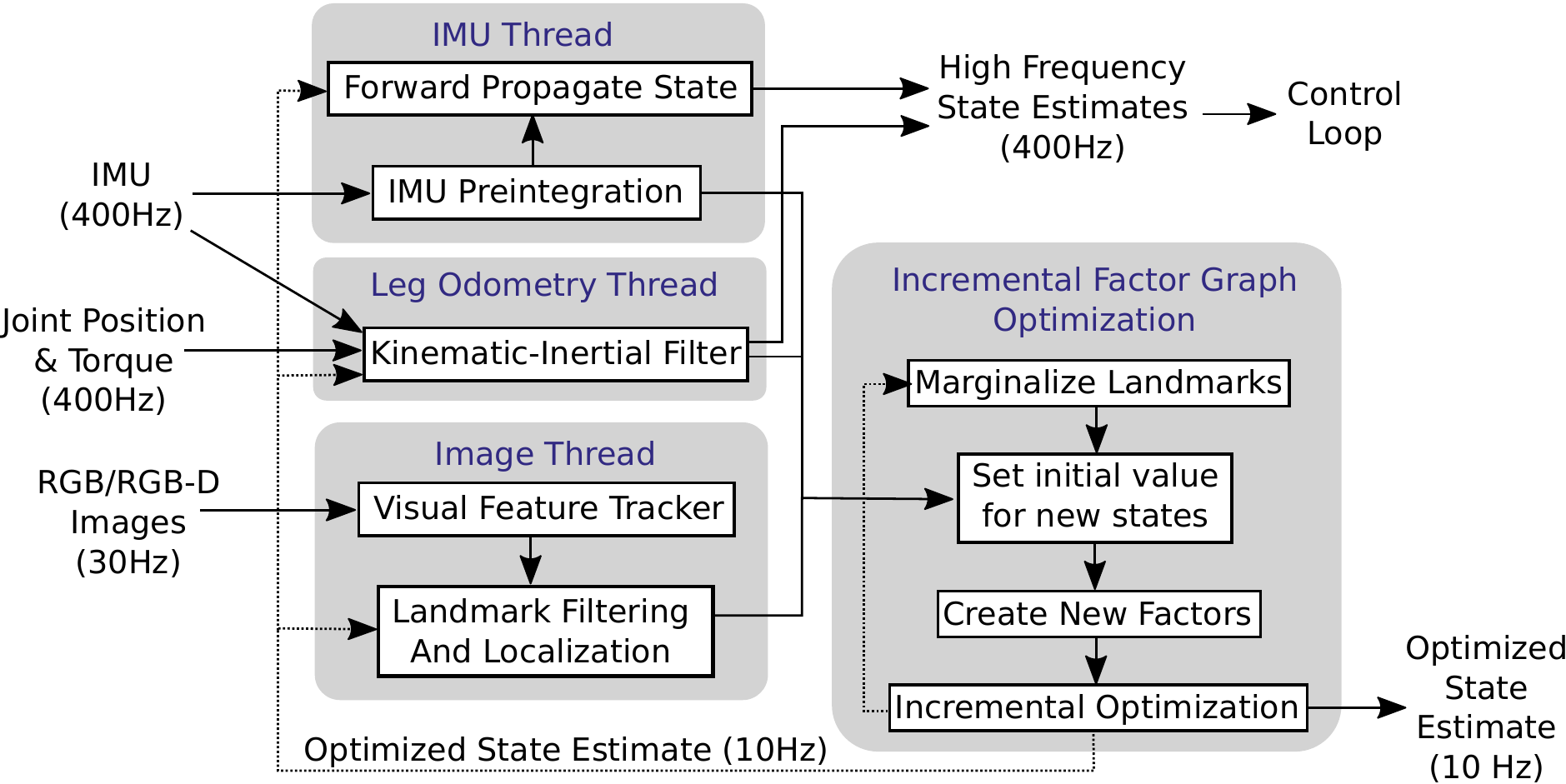}
\caption{The proposed state estimation architecture consists of measurement
handlers which process the sensor input and the current state from the optimizer to
create the factors. These factors are then smoothed using iSAM2 for incremental
factor graph optimization.}
\label{fig:algorithm-structure}
\vspace{-3mm}
\end{figure}

The factor graph optimization was implemented using the iSAM2 incremental
optimization library \cite{Kaess2012}. The structure of the system is shown in
\Figure \ref{fig:algorithm-structure}. The algorithm consists of a series of
measurement handlers running in separate threads that process the different
sensor inputs. When a new node is created (\eg when an image measurement is
received) each of the measurement handlers adds new factors to the graph. The
output of the factor graph optimization is then fed back to
the measurement handlers.
\subsection{Synchronization}
\label{sec:synch}
\begin{figure}
\vspace{3mm}
	\centering
	\includegraphics[width=0.9\columnwidth]{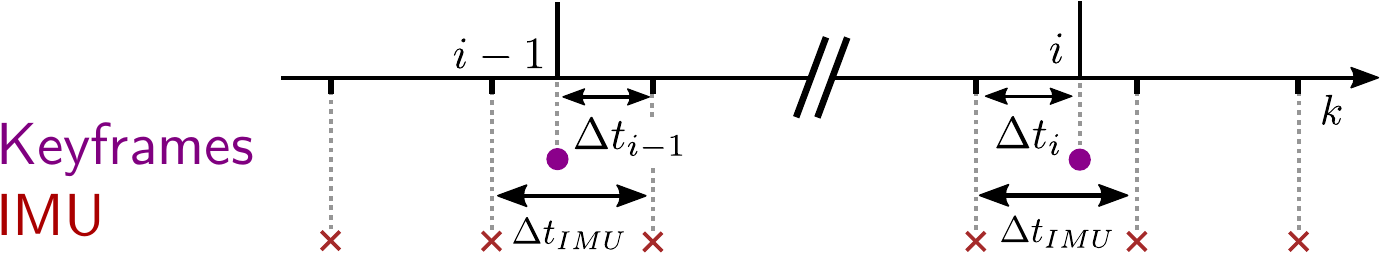}
	\caption{We consider and account for camera/IMU asynchronousity so as to improve the accuracy of the IMU factors. }
	\label{fig:imu-synchronization}
\vspace{-3mm}
\end{figure}

Given two consecutive keyframe indices $i-1,i \in \mathsf{K}_k$, preintegrating
all
IMU measurements between times $t_{i-1}$ and
$t_i$ would result in an incorrect motion estimation as the IMU measurement
timestamps may
not be aligned with the image frames (see \Figure \ref{fig:imu-synchronization}). To
avoid this, we correct the $\Delta t$ of the IMU measurements directly
before
and after the camera image timestamp $t_i$:
\begin{eqnarray}
\Delta t_{i-1} & = & t_\textsc{imu} + \Delta t_\textsc{imu} - t_i \\
\Delta t_{i}   & = & t_i - t_\textsc{imu}
\end{eqnarray}
where we assume constant acceleration and angular velocity between IMU
measurements.

\subsection{Visual Feature Tracking}

A core component of the VILENS system is the tight integration of visual
features into the optimization. It can provide lower drift state estimates by
tracking and optimizing the robot state and landmark positions over many
observations. Our visual feature tracking method is based upon the robust
pixel-based tracking approaches used in ROVIO \cite{Bloesch2017} and VINS-Mono
\cite{Qin2017}.

Visual features are first detected using the Harris Corner
Detector and then tracked through successive images using the
Kanade-Lucas-Tomasi feature tracker. This method provides sub-pixel accuracy
and is well-suited to the constrained but jerky motions
typical of a legged robot. After the tracking, outliers are rejected using
RANSAC with a fundamental matrix model similar to
\cite{Qin2017}. New features are then detected in the image to maintain a
minimum number of tracked features. These new features are constrained to be a
minimum distance (in image space) from existing features, to ensure an even
distribution of features across the image.

To limit the graph growth, we estimate the location of a feature in the world
frame only after it has been observed more than $N_\text{obs} = 30$ times. When
depth
is available, it is used for the initial landmark location estimate. When only
monocular data are available, we triangulate the landmark location
$\mathbf{m}_{j,0}$ using the last $N_\text{obs}$ frames with the Direct Linear
Transformation (DLT) algorithm from \cite{Carlone2014}. If the landmark is
successfully triangulated with a depth smaller than $d_\text{max} =
\SI{50}{\meter}$ then this and
successive measurements of the same landmark are added to the graph.

\subsection{Marginalization}
An important consideration in a legged robot system is latency, and without
marginalization, the time taken by iSAM2 optimization increases over time
\cite{Kaess2012}. This becomes an important consideration when operating the
robot for extended periods of time.

We marginalize states older than a threshold (typically,
\SI{10}{\second})
 and landmarks which are no longer observed, whilst keeping a minimum
number of nodes in the factor graph. These marginalized states and
landmarks are
replaced with a simple linear Gaussian factor based on the current linearization
point of the node. This Gaussian factor has the same form as the residual
defined in (\ref{eq:prior-residual}).


\section{Experimental Results}
\label{sec:experimental-results}
In this section, we present the experimental evaluation of the
proposed algorithm on three different datasets: EuRoC, Keble
College and Oil Rig. The first dataset is a purely VINS dataset
collected on a MAV. The other two datasets were collected using our ANYmal robot
in different outdoor environments: a college campus and an
industrial oil rig firefighter training facility.

\subsection{EuRoC Dataset}
\label{sec:euroc}
\begin{figure}
\vspace{3mm}
\centering
\includegraphics[height=6cm]{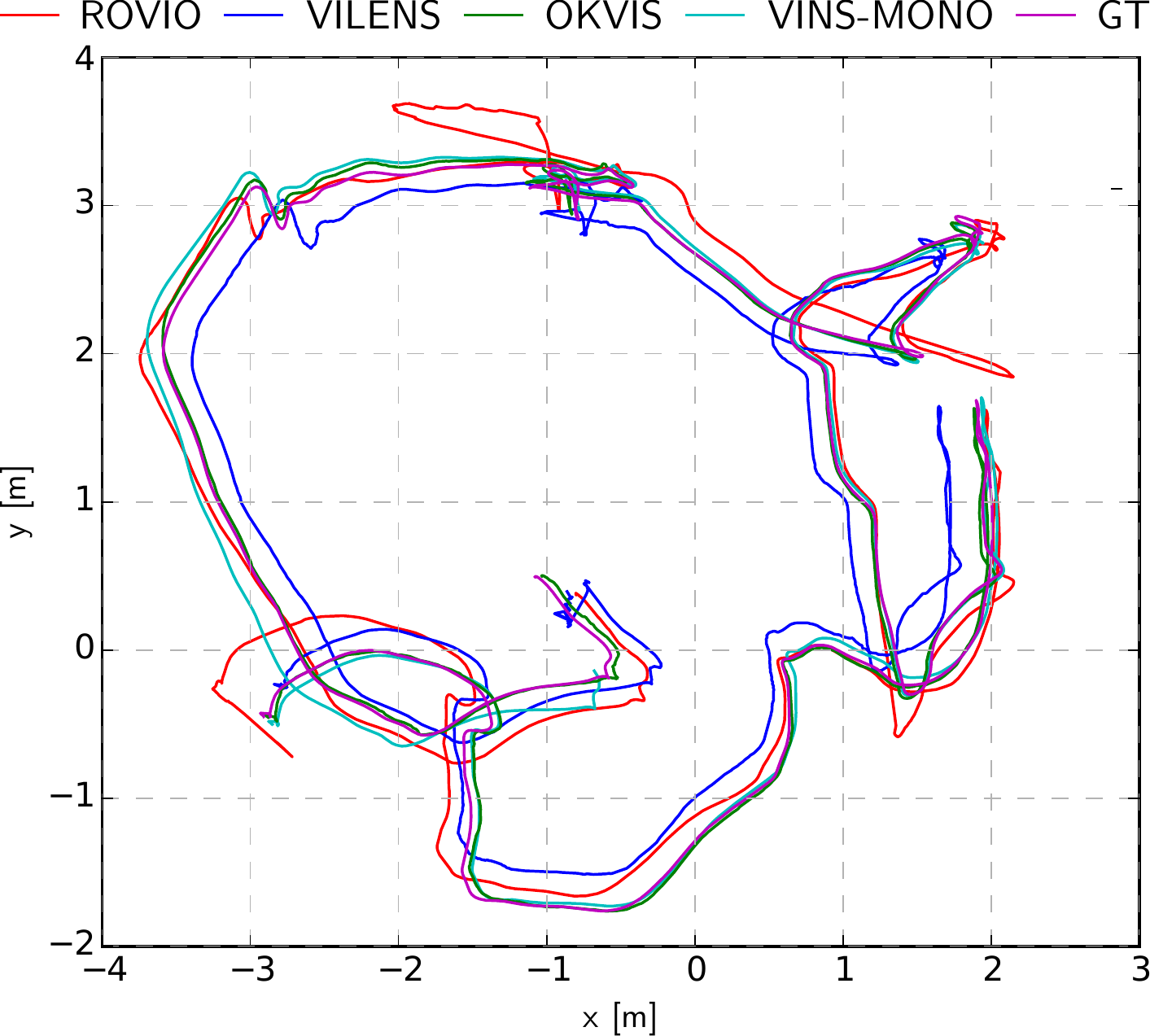}
\caption{Performance comparison between our baseline VILENS system (blue), ROVIO (red), OKVIS (green), and VINS-Mono
(cyan) compared to VICON ground truth for EuRoC V2\_01 dataset.}
\label{fig:euroc-v201-top-down}
\vspace{-3mm}
\end{figure}

To demonstrate that our approach builds upon a stand-alone VINS system, we
evaluated it on a portion of the \mbox{EuRoC} benchmark dataset
\cite{Burri2016} and compared it
to several state-of-the-art VINS algorithms including OKVIS
\cite{Leutenegger2015}, ROVIO \cite{Bloesch2017}, and VINS-Mono \cite{Qin2017}.
The estimated trajectory from the EuRoC V2\_01 dataset is shown in
\Figure \ref{fig:euroc-v201-top-down}.

In brief, we found that our core system can achieve
comparable performance to these VINS
algorithms. This
demonstrates that the estimator can function
without leg odometry information, which is important when that modality
becomes unreliable (\eg on slippery or soft ground).

\subsection{Outdoor Datasets Experimental Setup}
The outdoor experiments were conducted using the \mbox{12-DoF}
ANYmal quadruped
\cite{Hutter2016} (\Figure
\ref{fig:anymal-fsc}).

The measurements from the motors (\ie joint states) were synchronized via
EtherCAT, while the other sensors were synchronized by Network Time Protocol
(NTP). The sensor
configurations used for the two datasets are specified in Table
\ref{tab:sensors}.

To generate ground truth, we collected a dense
gravity-aligned prior map of the site using the Leica BLK-360 3D
laser scanner. Afterwards, we performed ICP localization (using the AICP
algorithm \cite{Nobili2017a}) against the prior map using data from ANYmal's Velodyne
VLP-16 LIDAR. Both map and LIDAR sensor were used as the ground truth only.
This provided a ground truth
trajectory with approximately \SI{5}{\centi\meter} accuracy, but only at
\SI{2}{\hertz}.

\subsection{Keble College Dataset}

\begin{figure}[t]
\vspace{3mm}
	\centering
	\includegraphics[height=5cm]{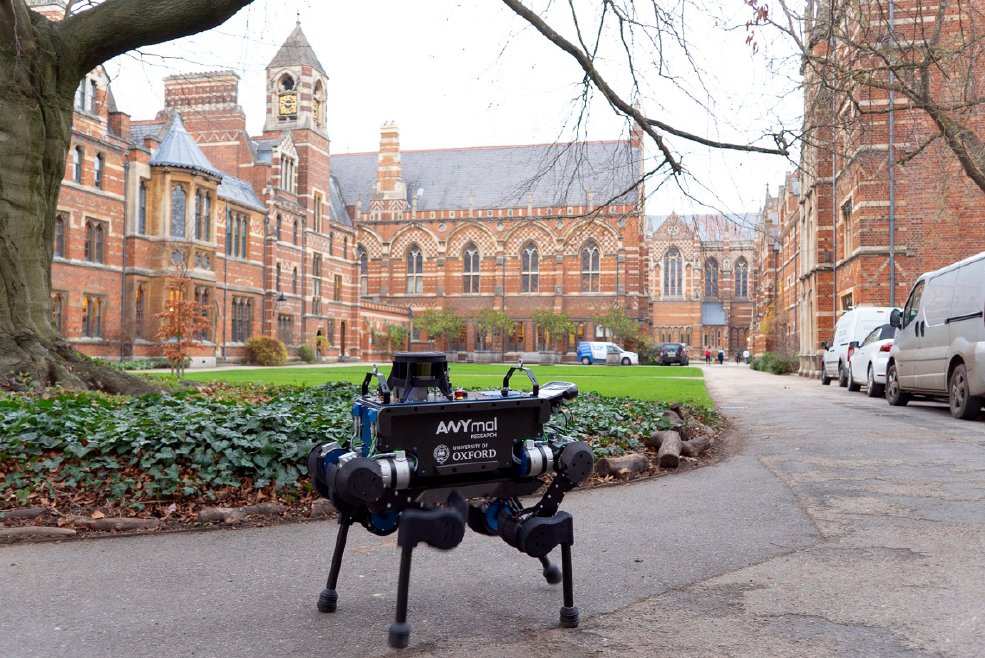}
	\caption{The Keble College dataset involved the ANYmal trotting around an
	open urban environment.}
	\label{fig:anymal-keble}
\end{figure}

The first outdoor dataset was collected in a urban environment at Keble
College, Oxford, UK. The dataset consists of the robot trotting on a concrete
path around a \SI{28 x 60}{\meter} open lawn surrounded by a
residential building (\Figure \ref{fig:anymal-keble}). The main challenges were
vegetation moving in the wind, long distances to visual features
(\SI{>10}{\meter}), and limited angular motion, which made feature triangulation
difficult.

We ran four trials each approximately \SI{22}{\meter} in length
and evaluated the mean and standard deviation of the Relative Position Error
(RPE) over a \SI{10}{\meter} distance (see Table \ref{tab:keble-results-table}).
Compared to the kinematic-inertial estimator, our algorithm
reduces the RPE by
\SI{15}{\percent} to \SI{55}{\percent} and yaw error by \SI{42}{\percent} to
\SI{85}{\percent},
as visual features tracked over many frames constrain pose drift.

\begin{table}
\centering
\caption{Mean (and standard deviation) performance on the Keble College
and Oil Rig datasets.}
\begin{tabular}{l|cc|cc}
\toprule
 & \multicolumn{2}{c|}{\textbf{RPE $\boldsymbol{\mu (\sigma)}$  [\si{\metre}]}}
& \multicolumn{2}{c}{\textbf{Yaw
Error $\boldsymbol{\mu (\sigma)}$ [\si{\deg}]}} \\
\midrule
\textbf{Dataset} & \textbf{TSIF} \cite{Bloesch2017b} & \textbf{VILENS} &
\textbf{TSIF} \cite{Bloesch2017b} &
 \textbf{VILENS} \\
\midrule
Keble 1 & 0.53 (0.21) & \textbf{0.30} (0.12) & 6.64 (2.23) & \textbf{0.99}
(0.80) \\
Keble 2 & 0.51 (0.10) & \textbf{0.23} (0.10) & 5.72 (0.94) & \textbf{1.47}
(1.07) \\
Keble 3 & 0.67 (0.10) & \textbf{0.52} (0.15) & 6.68 (0.80) & \textbf{3.86}
(1.90)\\
Keble 4 & 0.47 (0.11) & \textbf{0.40} (0.10) & 3.32 (1.15) & \textbf{1.13}
(1.46)\\
\midrule
Oil Rig & 0.44 (0.37) & \textbf{0.41} (0.18) & 4.89 (3.38) & \textbf{3.68}
(4.10) \\
\bottomrule
\end{tabular}
\label{tab:keble-results-table}
\vspace{-3mm}
\end{table}

\subsection{Oil Rig Dataset}
A second outdoor dataset was collected at an industrial firefighter training
facility in Moreton-In-Marsh, UK (\Figure
\ref{fig:anymal-fsc}). The facility closely matches the locations where ANYmal
is likely to be deployed in future.

\begin{figure}
\centering
\includegraphics[width=0.4\columnwidth]{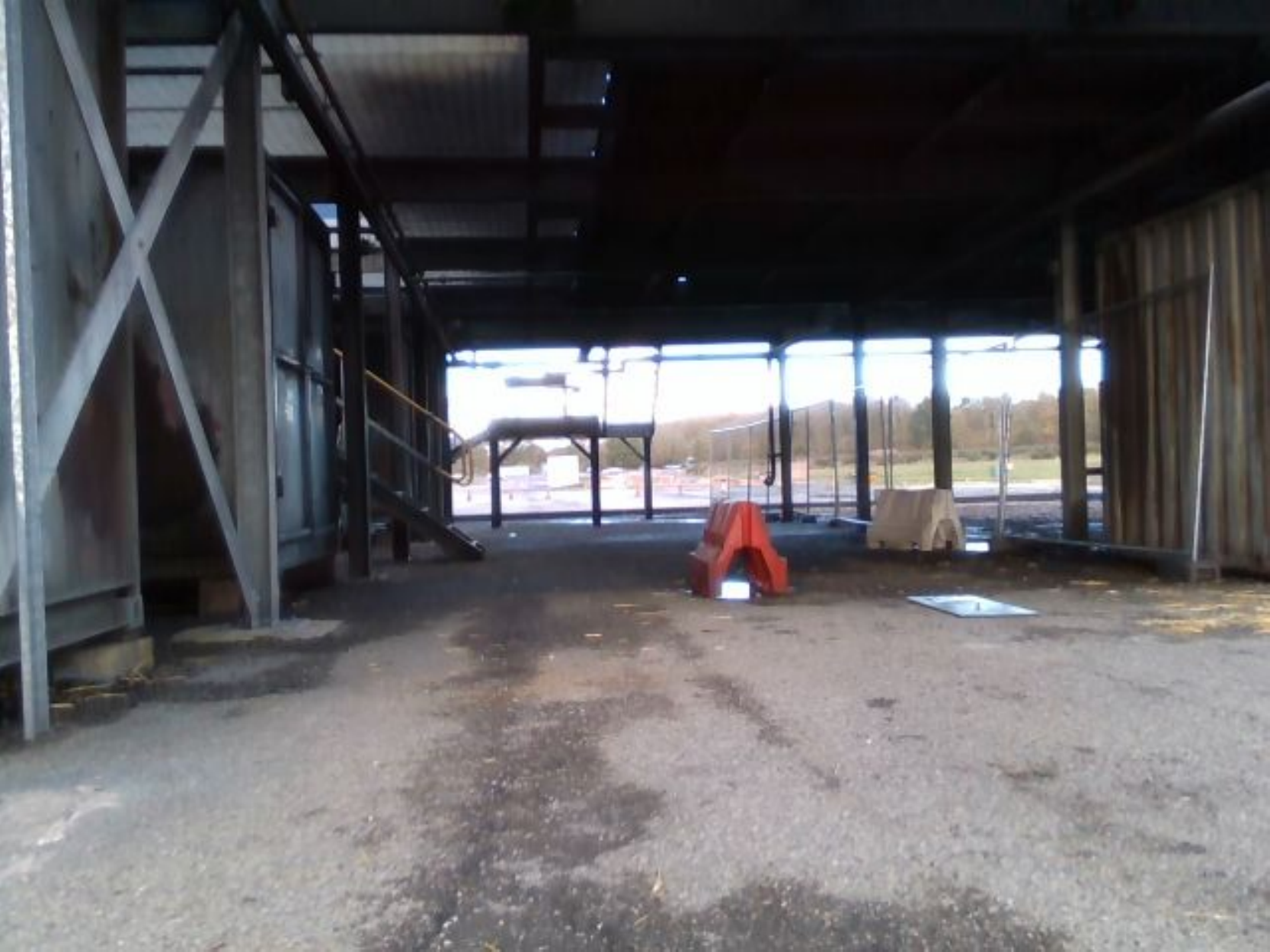}%
\hspace{0.01\columnwidth}%
\includegraphics[width=0.4\columnwidth]{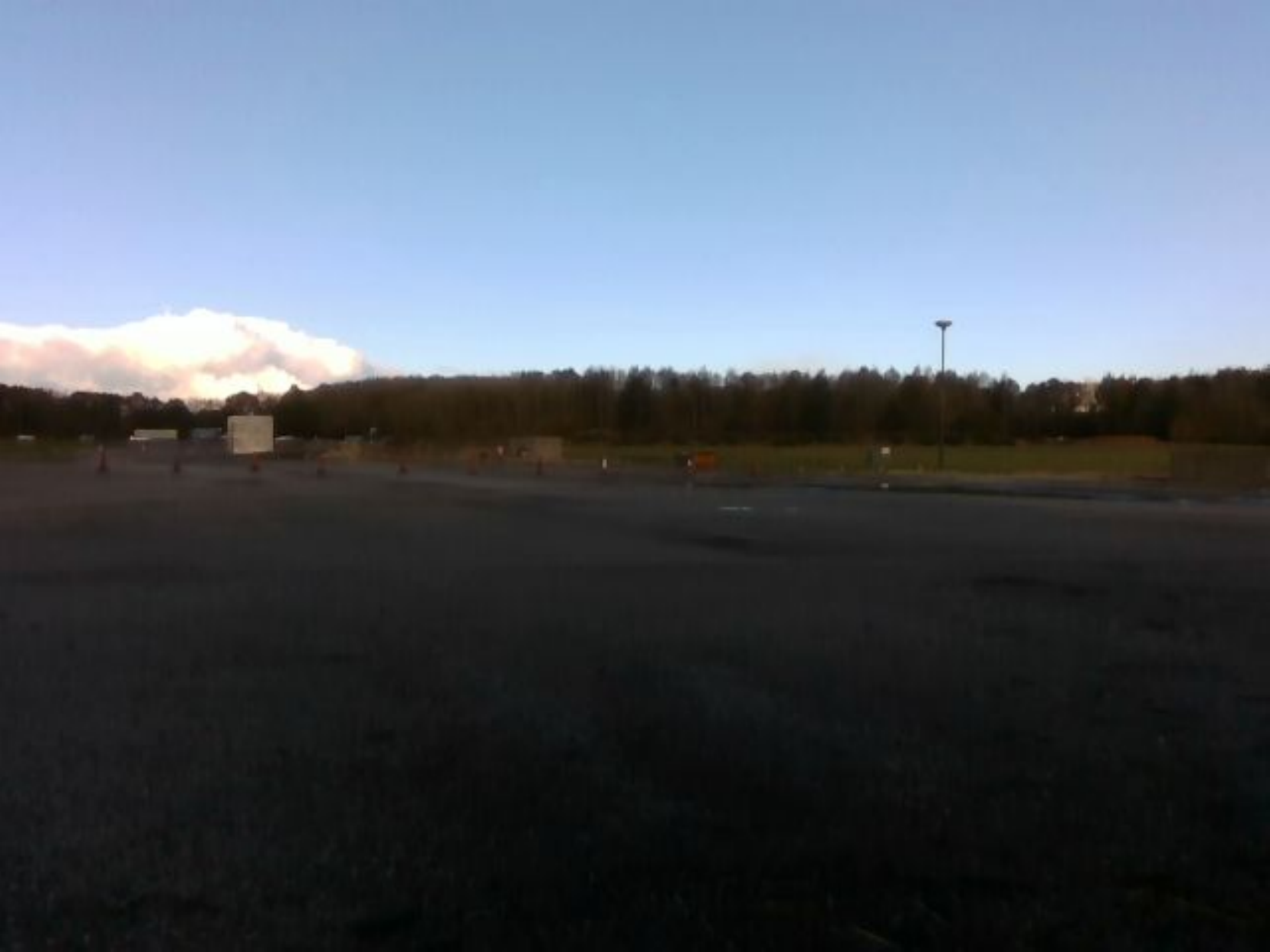}\\
\vspace{0.01\columnwidth}
\includegraphics[width=0.4\columnwidth]{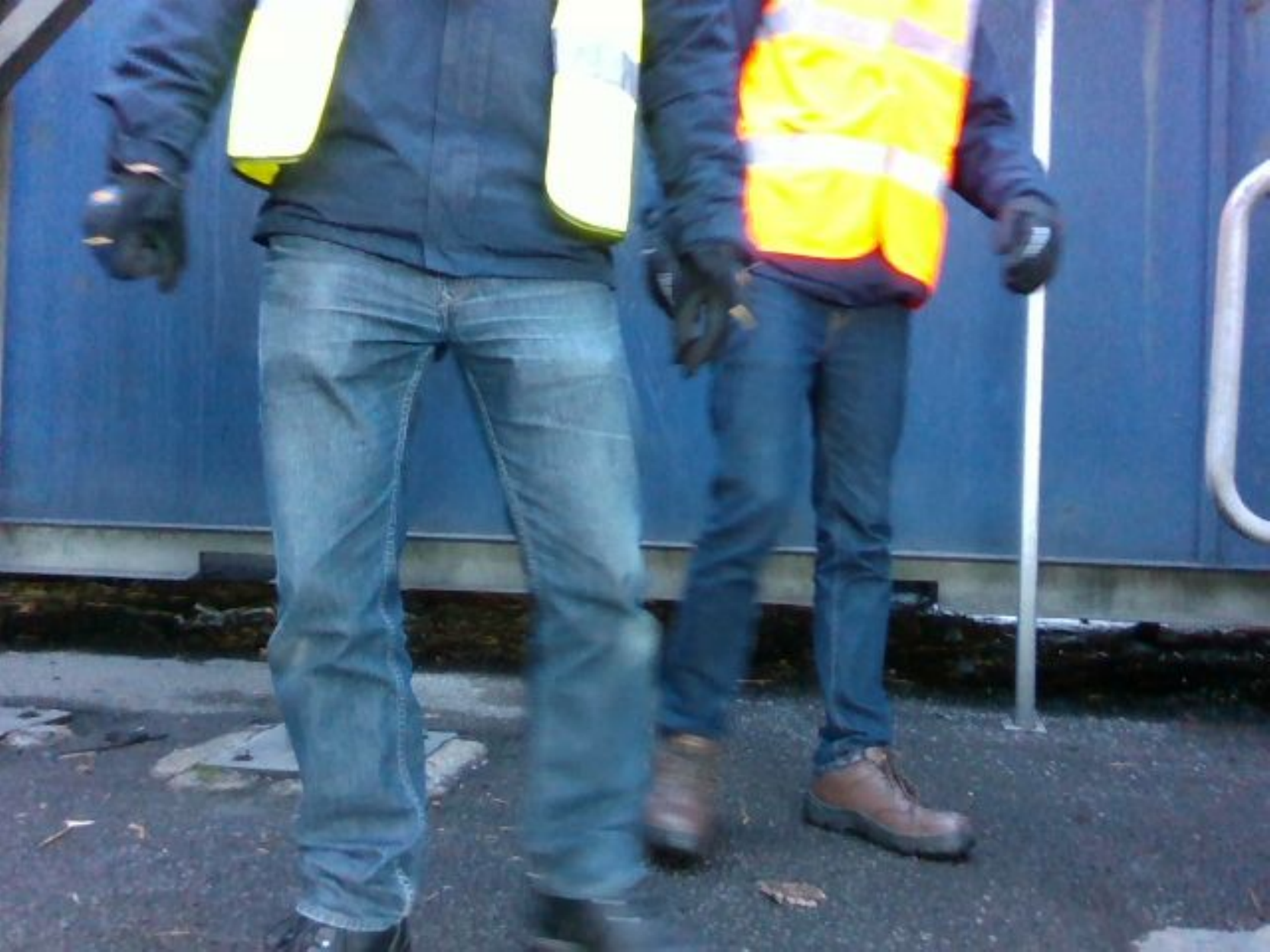}%
\hspace{0.01\columnwidth}%
\includegraphics[width=0.4\columnwidth]{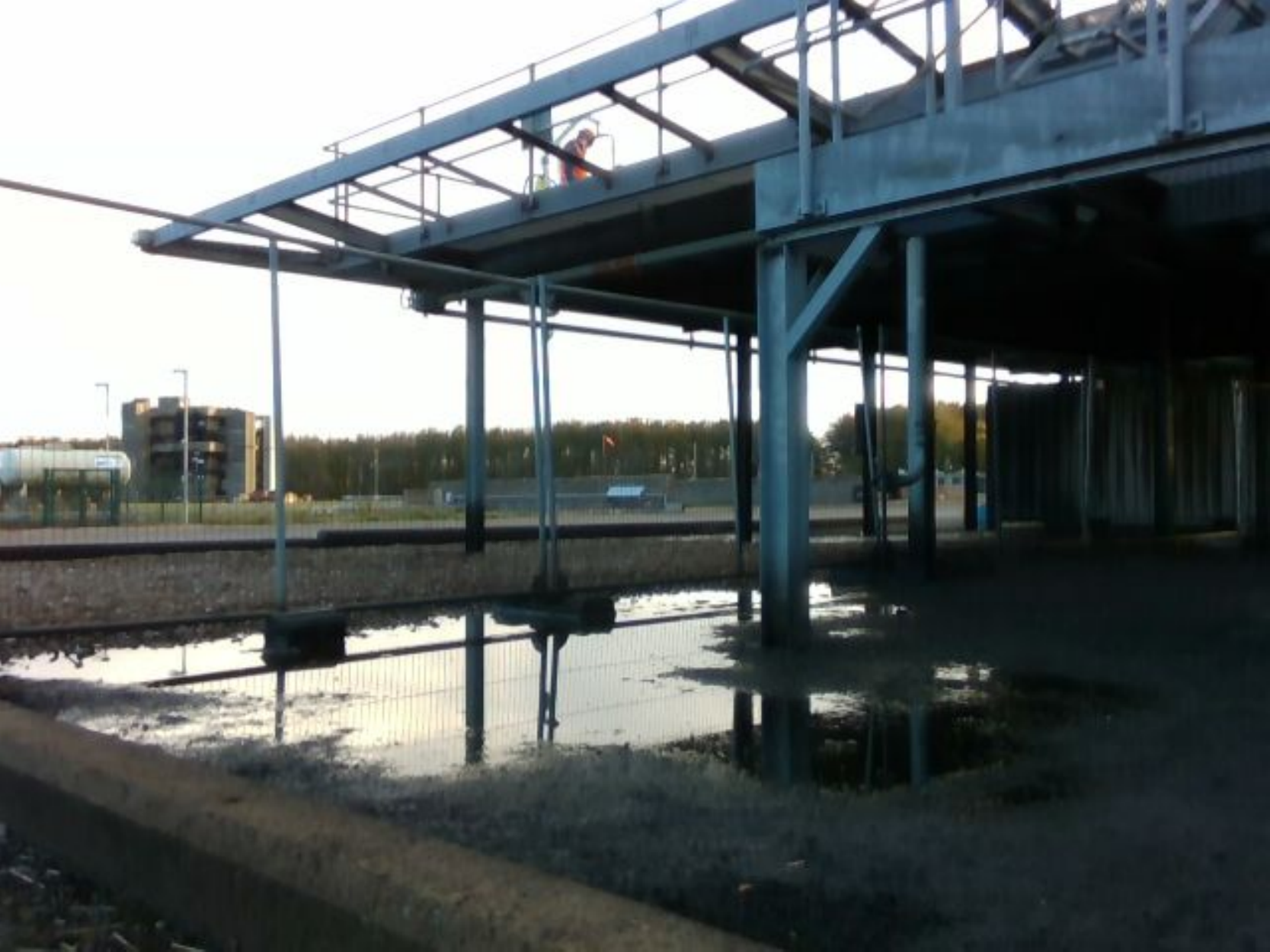}
\caption{Notable situations within the Oil Rig dataset. \emph{Top-Left:} VILENS
outperforms TSIF where
there are many visual features. \emph{Top-Right:} VILENS and TSIF perform
similarly where
there is no structure in front of the robot. \emph{Bottom:}
Moving objects and reflections are robustly handled by VILENS.}
\label{fig:fsc-scenarios}
\vspace{-6mm}
\end{figure}

This \SI{110}{\meter} (\SI{22}{\minute}) long dataset involves the ANYmal
robot trotting through the facility, climbing over a
slab, and walking up a staircase into a smoke-blackened room. Challenging situations
include: featureless areas,
stationary periods with intermittent motion, dynamic obstacles
occluding large portions of the image, non-flat terrain traversal, and foot
slip caused by a combination of mud, oil, and water on the ground (\Figure
\ref{fig:fsc-scenarios}).

\begin{figure}
\vspace{3mm}
\centering
\includegraphics[width=\columnwidth]{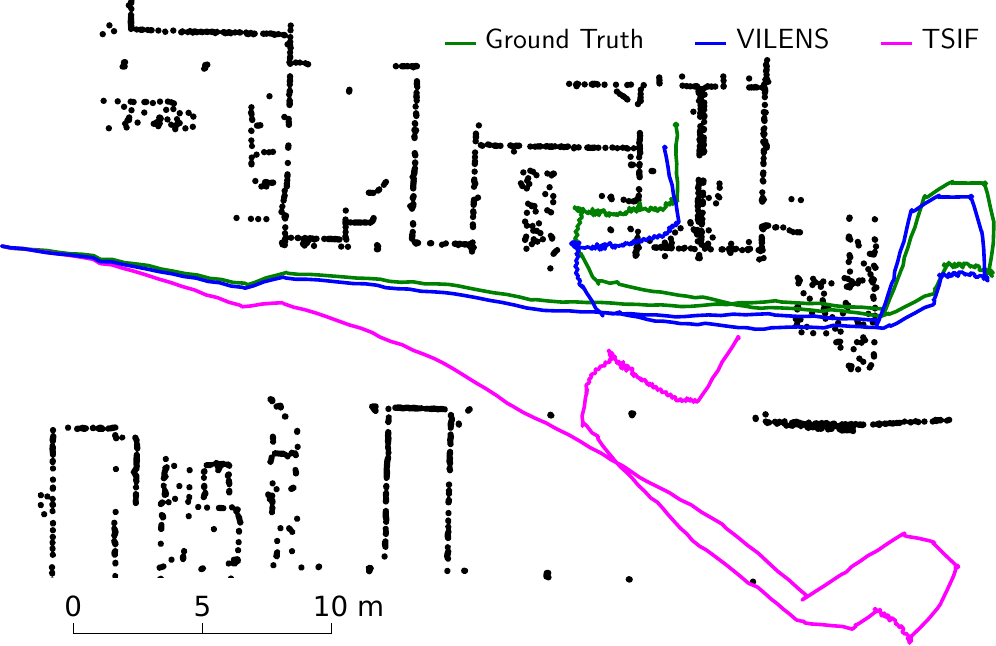}
\caption{Top view of the estimated trajectories of VILENS (blue), TSIF
(magenta), and ground truth (green) on the Oil Rig Dataset. The
map is not used in the algorithm and is shown for illustrative purposes only.}
\label{fig:fsc-trajectory}
\vspace{-3mm}
\end{figure}

Figure \ref{fig:fsc-trajectory} shows the estimated trajectory from VILENS,
compared to TSIF \cite{Bloesch2017b} and ground truth. The Absolute Translation
Error (ATE) for VILENS is \SI{76}{\percent}
lower compared to the TSIF (\SI{1.65}{\metre} and \SI{6.88}{\metre},
respectively).

Looking at the relative performance over \SI{10}{\meter}, the error reduction is
\SI{7}{\percent} for RPE and \SI{25}{\percent} for yaw (Table
\ref{tab:vins-results-table}). This suggests that the Oil Rig dataset is more
challenging than Keble, since the accuracy at small scale is closer to the TSIF
(see Section \ref{sec:discussion-and-limitations}).

Note that performance evaluation against the VINS algorithms mentioned in
Section~\ref{sec:euroc} was not possible because they either failed to
initialize due to the lack of motion or diverged after a short period.
This is also
confirmed in Table \ref{tab:vins-results-table}, where we evaluated the
performance of VILENS as standalone VINS system against TSIF and full VILENS.
Without Leg
Odometry factors, VILENS fails for all the datasets except the first
\SI{70}{\second} of Keble 1 and Oil Rig, where it performs worse or same.

\begin{table}
\centering
\caption{Mean (and standard deviation) performance of TSIF,
VILENS as VINS system, and VILENS before VINS failure (\SI{\sim70}{\second})}
\begin{tabular}{l|ccc|ccc}
\toprule
& \multicolumn{3}{c|}{\textbf{RPE $\boldsymbol{\mu (\sigma)}$  [\si{\metre}]}}
& \multicolumn{3}{c}{\textbf{Yaw
Error $\boldsymbol{\mu (\sigma)}$ [\si{\deg}]}} \\
\midrule
\textbf{Dataset} & \textbf{TSIF} & \textbf{VINS} & \textbf{VILENS} &
\textbf{TSIF} & \textbf{VINS} & \textbf{VILENS} \\
\midrule
\multirow{2}{*}{Keble 1}
& 0.30   & 0.36   & \textbf{0.25} & 4.28   & \textbf{0.74} & 0.75 \\
& (0.06) & (0.12) & (0.13)        & (1.07) & (0.61) & (0.56) \\
\midrule
\multirow{2}{*}{Oil Rig}
& 1.09   & 5.33   & \textbf{0.34} & 10.38  & 5.03   & \textbf{1.21} \\
& (0.09) & (0.53) & (0.12)        & (0.75) & (3.21) & (0.90) \\
\bottomrule
\end{tabular}
\label{tab:vins-results-table}
\vspace{-3mm}
\end{table}

\subsection{Timing}

A summary of the key computation times in the proposed algorithm are shown in
Table \ref{tab:timing-table}. Since the kinematic-inertial filter, the
image processing, and the optimization are run in separate threads, the system
is capable of outputting a high-frequency kinematic-inertial state estimate
(for control purposes) at \SI{400}{\hertz}, and an optimized estimate
incorporating visual
features at approximately \SI{10}{\hertz}.

\begin{table}
\centering
\caption{Mean (and standard deviation) processing time for components of the
VILENS system, on the Oil Rig dataset.}
\begin{tabular}{llr}
\toprule
{\textbf{Thread}} & {\textbf{Module}}  &
$\boldsymbol{\mu (\sigma)}$ \textbf{[\si{\milli\second}]}\\
\midrule
\multirow{3}{*}{Optimization} & Factor Creation & 10.80 (4.50) \\
                              & Optimization    & 10.05 (7.69) \\
                              & Marginalization & 0.82  (0.97) \\
                              \midrule
                              & \textbf{Total}  & \textbf{21.67}
(\textbf{13.12}) \\
\midrule
\multirow{4}{*}{Image Proc}  & Image Equalization & 0.87  (0.51) \\
                             & Feature Tracking   & 2.04  (0.98)  \\
                             & Outlier Rejection  & 1.74  (1.87) \\
                             & Feature Detection  & 5.33  (0.91) \\
                             \midrule
                             & \textbf{Total}     & \textbf{9.99}
(\textbf{4.27}) \\
\bottomrule
\end{tabular}
\label{tab:timing-table}
\vspace{-3mm}
\end{table}

\section{Discussion}
\label{sec:discussion-and-limitations}
In the previous section, we have demonstrated that VILENS outperforms
kinematic-inertial and visual-inertial methods for all the datasets.
However, from Table \ref{tab:keble-results-table} we can see that the gap
between VILENS and TSIF is not uniform across the datasets.

An in-depth analysis of performance was limited by the accuracy and frequency
of our
LIDAR ground truth. Nonetheless, we see that the feature quality has an influence of the
drift rate over small scales.

\begin{figure}
\vspace{3mm}
\centering
\includegraphics[width=\columnwidth]{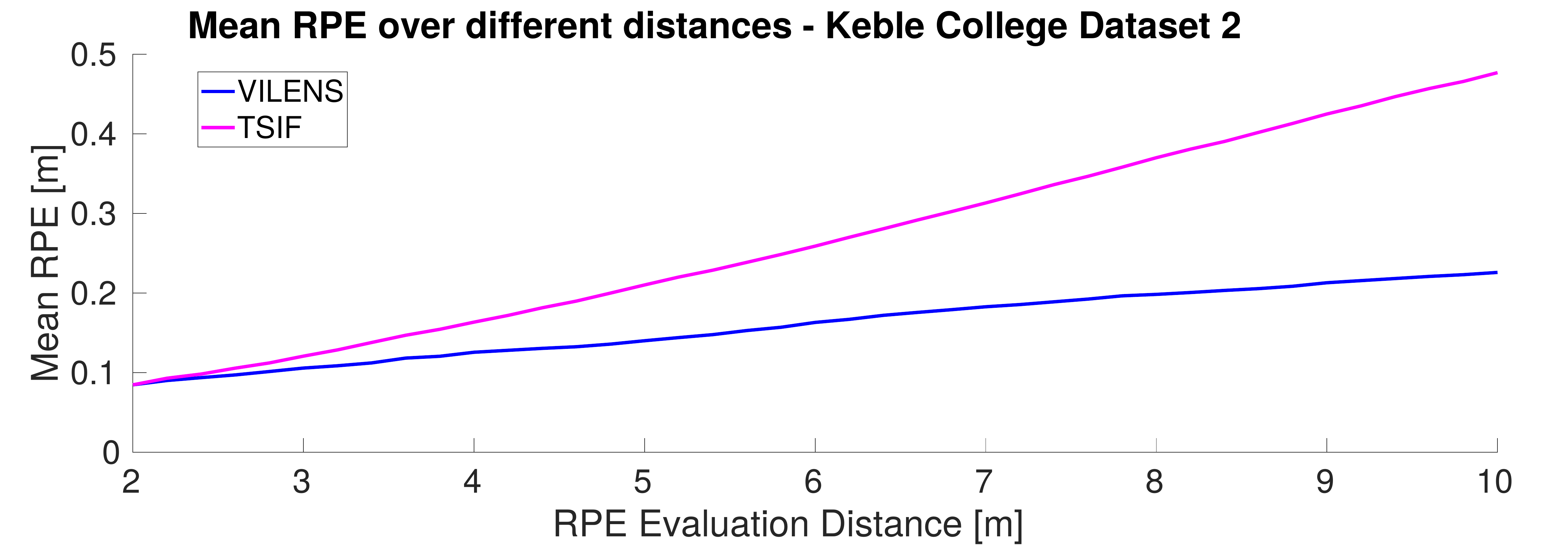} \\
\includegraphics[width=\columnwidth]{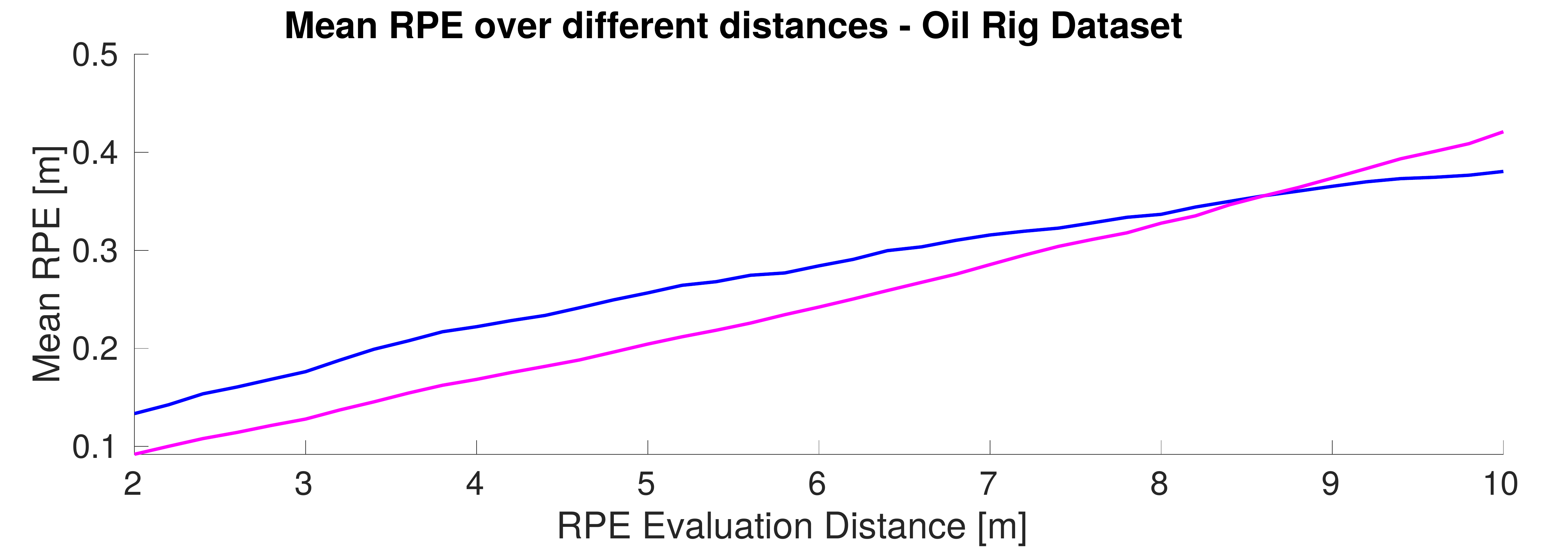}%
\caption{Mean RPE for VILENS and TSIF at different distance scales.
\emph{Top:} Keble College Dataset. \emph{Bottom:} Oil Rig Dataset. The Oil Rig
dataset is more challenging and therefore VILENS outperforms TSIF at
larger distance (\SI{~8.5}{\meter}) than for Keble.}
\label{fig:rpe-dist}
\vspace{-6mm}
\end{figure}

In \Figure \ref{fig:rpe-dist} we compare the drift rate of VILENS
and TSIF over different distance scales for the Keble College and
the Oil Rig datasets. The latter experiment is more challenging from the
perspective of exteroception and as a result the scale at which VILENS starts
to outperform TSIF
is larger. This is due to the tracking of poor quality features in the conditions
highlighted in \Figure \ref{fig:fsc-scenarios}.

In future work, we are motivated to improve performance in these challenging scenarios by exploring the use
of redundant or wider field-of-view cameras as well as different methods of
incorporating the kinematics directly into the factor graph.
Additionally, as part of an ongoing project, we intend to test performance in soft and compliant surface materials such
as sand, mud and gravel where we envisage the visual part of the estimator
predominating during sinking, sliding and slipping.

\section{Conclusion}
\label{sec:conclusions}
In this paper, we have presented VILENS (Visual Inertial LEgged Navigation
System), a robust state estimation method for legged
robots based on factor graphs, which
incorporates kinematic, inertial and visual information. This method outperforms
the robot's kinematic-inertial estimator and robustly estimates the robot
trajectory in
challenging scenarios, including textureless areas, moving occludants,
reflections and slippery ground. Under the same conditions,
current state-of-the-art visual-inertial algorithms diverge rapidly.



\bibliographystyle{./IEEEtran}
\bibliography{./IEEEabrv,./library}

\end{document}